\documentclass{article}

\usepackage{nac_preprint}

\usepackage[utf8]{inputenc} 
\usepackage[T1]{fontenc}    
\usepackage{hyperref}       
\usepackage{url}            
\usepackage{booktabs}       
\usepackage{amsfonts}       
\usepackage{nicefrac}       
\usepackage{microtype}      

\usepackage{booktabs} 
\usepackage{subcaption}
\usepackage{amsmath}
\usepackage{mathtools}
\usepackage[toc,page]{appendix}
\usepackage{enumitem}
\usepackage{graphics}
\usepackage{graphicx}
\usepackage{algorithm}
\usepackage{algorithmicx}
\usepackage[noend]{algpseudocode}
\algnewcommand{\LineComment}[1]{\State \(//\) #1}

\title{Towards a Predictive Processing Implementation of the Common Model of Cognition}

\author{%
  Alexander Ororbia \\
  Rochester Institute of Technology \\
  Rochester, NY, USA \\
  \texttt{ago@cs.rit.edu}\\
  \And
  M. Alex Kelly \\
  Bucknell University,
  Lewisburg, PA, USA\\
  Carleton University, Ottawa, ON, Canada\\
  \texttt{AlexKelly@cunet.carleton.ca}
}

\begin{document}
\setlength{\abovedisplayskip}{0.065cm}
\setlength{\belowdisplayskip}{0pt}

\maketitle

\begin{abstract}
\label{sec:abstract}
In this article, we present a cognitive architecture that is built from powerful yet simple neural models. Specifically, we describe an implementation of the common model of cognition grounded in neural generative coding and holographic associative memory.
The proposed system creates the groundwork for developing agents that learn continually from diverse tasks as well as model human performance at larger scales than what is possible with existant cognitive architectures.

\end{abstract}	

\keywords{cognitive architecture \and predictive processing \and holographic memory \and continual learning}

\section{Introduction}
\label{sec:intro}

Modern machine learning techniques based on artificial neural networks (ANNs) are implemented through algebraic manipulations of vectors, matrices, and tensors in high-dimensional spaces. While ANNs have an impressive ability to process data to find patterns, they do not typically model high-level cognition. Furthermore, ANNs are usually models of only a single task. Otherwise, when an ANN is trained to learn a series of tasks, catastrophic interference occurs, with each new task causing the ANN to forget all previously learned tasks \cite{french1999catastrophic,mannering2021catastrophic,mccloskey_catastrophic_1989}. On the other hand, symbolic cognitive architectures, such as the widely used ACT-R \cite{anderson2009,Ritter2019actr},
can capture the complexities of high-level cognition but scale poorly to the naturalistic, non-symbolic data of sensory perception, e.g., images, or to big data sets necessary for modelling learning over a lifetime, e.g., corpora with hundreds of millions of words. 

Are symbolic and ANN approaches compatible? Symbolic and neural models can be understood as theories of cognition operating at different levels of description \cite{Kersten2016}. Is it possible to provide a theory that bridges these two levels, a reduction of the symbolic to the neural, while retaining the strengths and capabilities of each?

We propose a cognitive architecture that is built on two biologically plausible, neural models, namely a variant of predictive processing
known as Neural Generative Coding (NGC) \cite{ororbia2020continual} and holographic memory \cite{Kelly2020hdm}.
Desirably, the use of these particular building blocks yields naturally scalable, local update rules (based on variants of Hebbian learning \cite{hebb1949organization}) to adjust the overall system's synaptic weight parameters while facilitating robustness in acquiring, storing, and composing distributed representations of tasks encountered sequentially.
Our intent is to advance towards a cognitive architecture capable of modeling human performance at all scales of learning, from the half-hour lab experiment to skills acquired gradually over a lifetime. By combining predictive processing with vector-symbolic models of human memory, we create a model of cognition that has the power of modern machine learning techniques while retaining long-term memory, single-trial learning, transfer-learning, and other cognitive capacities associated with high-level cognition.

\begin{figure}
\begin{center}
\includegraphics[width=0.6\linewidth]{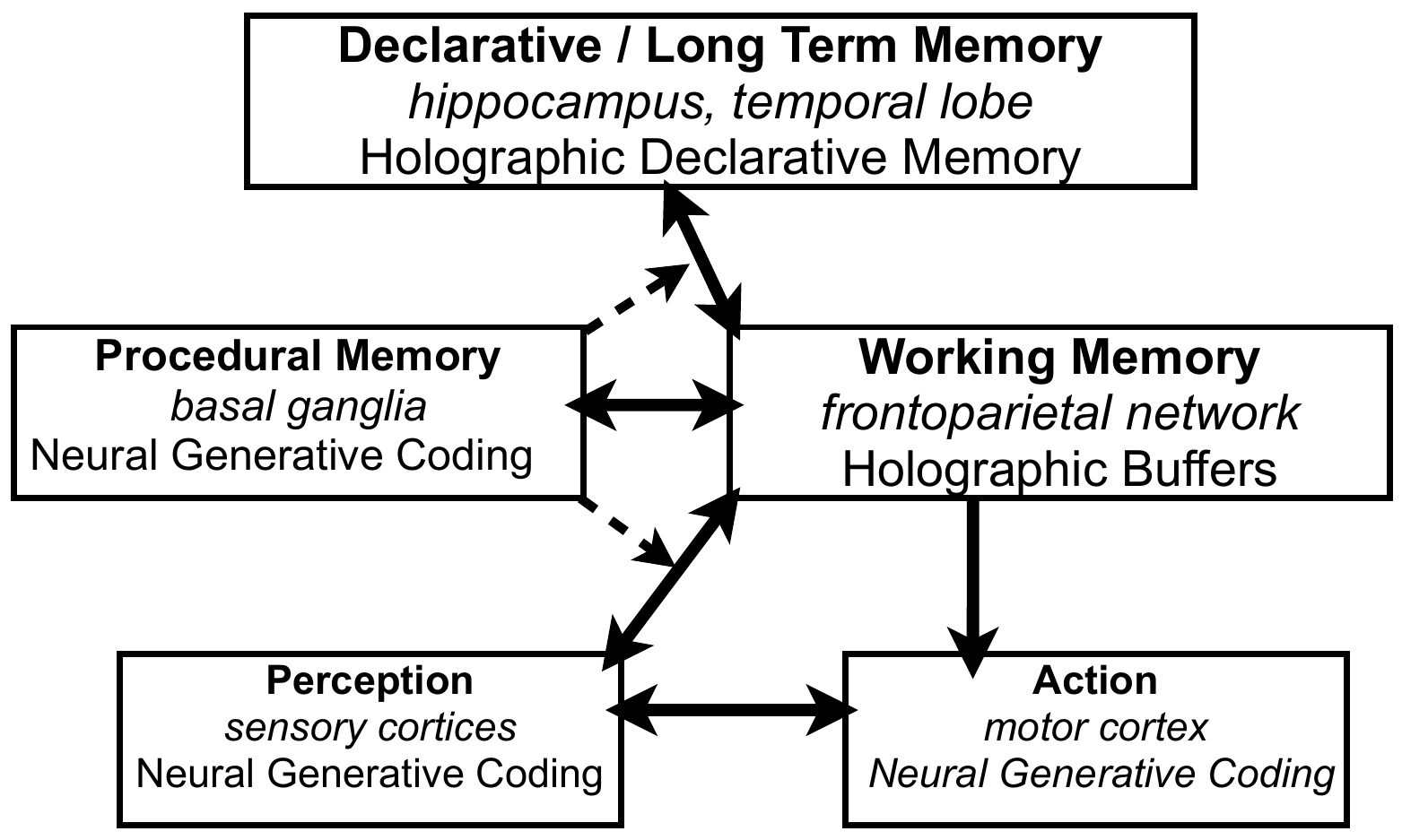}
\end{center}
\caption{Common Model of Cognition \protect\cite{laird2017standard}, associated brain areas \protect\cite{Stocco2021cmc,Stein-Hanson2018,Stocco2018}
and our approach to modelling each module. Solid arrows are data passing. Dashed arrows are modulation of data passing. 
}
\label{FigCMC}
\vspace{-0.5cm}
\end{figure}

\section{The Common Model of Cognition} 
\label{sec:cmc}

Since Newell \cite{Newell1973} first argued that good empirical work and piecemeal theoretical work are insufficient to understand the mind, researchers in cognitive science have sought to develop functional, testable theories of cognition as a whole. Cognitive architectures serve as both unified theories of cognition and as computational frameworks for implementing models of specific experimental tasks. Forty years of research has developed hundreds of cognitive architectures, many with strong similarities to each other \cite{Kotseruba2018}, suggesting an emerging consensus on the basic principles of cognition. In \cite{laird2017standard}, commonalities are found between three cognitive architectures, ACT-R \cite{Anderson1998}, Soar \cite{Laird2012}, and SIGMA \cite{Rosenbloom2016}. 
On the basis of the similarities, Laird, Lebiere, and Rosenbloom \cite{laird2017standard} propose a \emph{Common Model of Cognition}, a high-level theory of the modules of the mind and how they interact (see Fig. \ref{FigCMC}).

The Common Model of Cognition consists of perceptual and motor modules that interact with the agent's environment, working memory buffers which hold the active data in the agent's mind, a declarative or long-term memory module that holds the agent's world knowledge, and a procedural memory module that controls the flow of information and evaluates possible actions \cite{laird2017standard}. An evaluation of fMRI data from 200 participants across seven different tasks found correlations in patterns of activity across brain areas consistent with the Common Model of Cognition's modules and their interactions \cite{Stocco2021cmc}.

\section{Neural Generative Coding (NGC)}
\label{sec:ngc}

Neural Generative Coding (NGC) is an instantiation of the predictive processing brain theory \cite{rao1999predictive,friston2005theory,clark2015surfing}, yielding an efficient and robust form of predict-then-correct learning and inference. We describe how neural circuits taken from NGC framework can be used to design several key modules of our cognitive architecture.

\subsection{Neural Generative Sensory Cortices}
NGC can be used to design neural generative models responsible for processing a specific modality of data (e.g., image pixels or speech waveforms). In \cite{ororbia2020neural,ororbia2020continual}, we show that NGC learns a good density estimator of data (from which new samples can be sampled or ``fantasized''), in conjunction with desired target functionality (e.g., classification, regression), in not only the cases of static input but also in cases of time-varying data series. This ability of the NGC process modalities over time will form the sensory cortices of our architecture (see Fig. \ref{FigCortex}).

\subsection{Neural Generative Motor Cortex}
Another key element of a functioning agent is one that demonstrates impetus, i.e., the agent must not merely react to its environment but must also manipulate it. In line with this particular goal, the agent needs actuators that are driven by neural circuits that exhibit motor control. 
In \cite{ororbia2021adapting}, we generalize NGC to the case of action-driven tasks, common in reinforcement learning, which we call \emph{active neural generative coding} (ANGC) given that it builds upon concept of planning-as-inference \cite{botvinick2012planning}. This particular work provides promising evidence that NGC can be used to build a coupled generative model and controller system that is capable of solving important reinforcement learning problems, particularly those when the reward signal is sparse, or even non-existent. This model would serve as the motor cortex in our architecture (see Fig. \ref{FigCortex}).

\subsection{Neural Generative Basal Ganglia}
\begin{figure*}[t!]
    \centering
    \begin{subfigure}[t]{0.475\textwidth}
        \centering
        \includegraphics[height=2.25in]{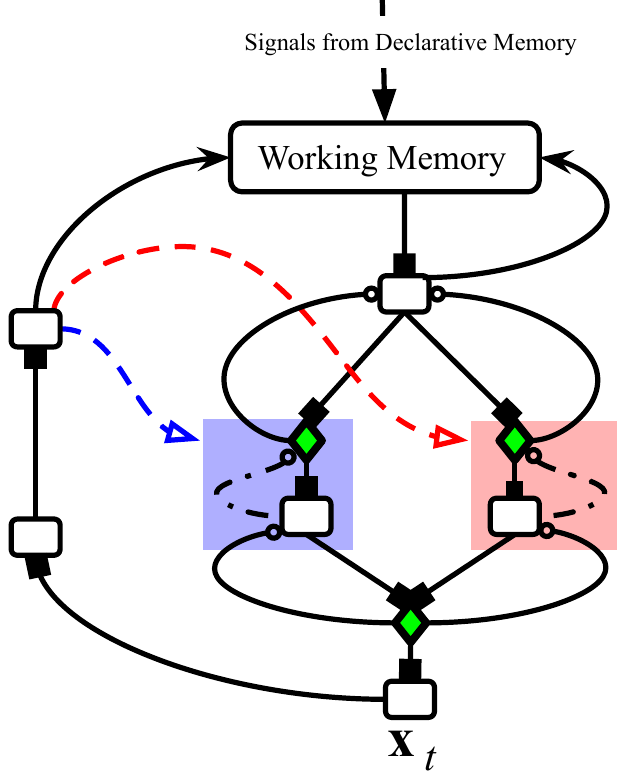}
        \caption{The basal ganglia, working memory, and sensory cortex sub-system working to reconstruct sensory input $\mathbf{x}_t$.}
        \label{FigGanglia}
    \end{subfigure}%
    \hspace{0.255cm} 
    \begin{subfigure}[t]{0.475\textwidth}
        \centering
        \includegraphics[height=1.85in]{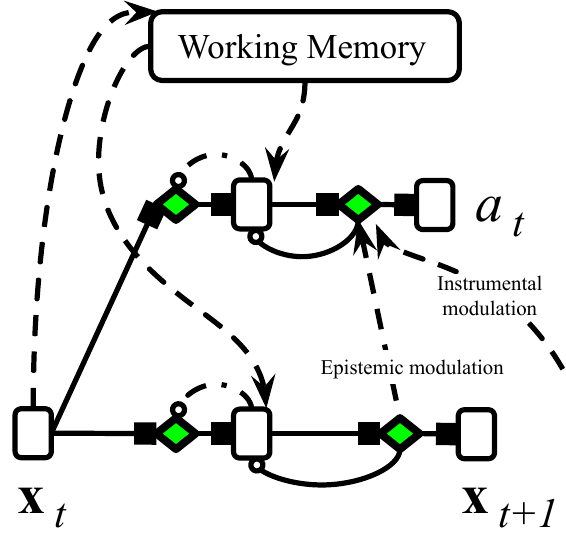}
        \caption{The working memory, perceptual cortex, and motor cortex sub-system working to product discrete action $a_t$.}
        \label{FigCortex}
    \end{subfigure}
    \caption{Potential implementations of two neural sub-systems within our instantiation of the Common Model of Cognition. Depicted are information loops that facilitate interfacing with working memory. Solid lines depict the transfer of information through either excitatory (lines with filled squares) or inhibitory (lines with non-filled circles) synapses while dashed lines depict signals that might be carried via synapses or transmitted via other mechanisms (for example, the blue and red dashed curves emanating from the basal ganglia would yield signals that multiplicatively gate the activities in the blue and red boxes respectively). 
    In the first diagram (Left), the basal ganglia model is shown inhibiting/suppressing the activities of one part of the sensory cortex (red box) while permitting another region to play a role in processing the input (blue box). In the second diagram (Right), the motor cortex, which is responsible for choosing actions, is depicted interacting with a sensory cortex -- the motor cortex adjust synapses use scalar ``dopamine'' signals driven by the environment (an instrumental signal to facilitate goal-oriented behavior) as well as from the sensory cortex (an epistemic signal to encourage exploration) \cite{ororbia2021adapting}.}
    \vspace{-0.5cm}
\end{figure*}

In \cite{ororbia2019lifelong}, we model the functionality of the basal ganglia in suppressing/inhibiting neural activity for the purpose of action selection and task switching \cite{yehene2008basal,cameron2010executive,stewart2012learning}, a behavior we argue is critical in facilitating effective continual learning without catastrophic interference.
The task selection model, which learns through competitive Hebbian learning, will serve as the basis for part of the basal ganglia in our cognitive architecture, acting to coordinate the exchange of information between the working memory and the sensory and long-term memory cortices (see Fig.~\ref{FigGanglia}). Note, as also depicted in Fig.~\ref{FigGanglia} and Fig.~\ref{FigCMC}, the basal ganglia is responsible for more than just interaction with the sensory cortices -- it is meant to interact with the working memory as well as modulate the transfer of information between the working memory and long term memory. 

\section{Holographic Memory}
\label{sec:holo_mem}

Holographic memory \cite{plate1995holographic}, also known as vector-symbolic architectures \cite{Gayler2003} or hyper-dimensional computing \cite{Kanerva2009}, is a formalism for capturing the capacity for humans to learn and recall arbitrarily complex associations between stimuli in the environment. Holographic memory is immune to the catastrophic interference typical of more conventional neural networks \cite{mannering2021catastrophic} allowing holographic memory to be used to create models that handle multiple, unrelated tasks \cite{Cheung2019superposition}. We propose to use holographic memory to implement the long-term declarative memory and the working memory buffers.

\subsection{Working Memory Buffers}
We implement a working memory buffer as a single holographic vector. Holographic memory vectors have a long established ability to account for benchmark memory phenomena such as serial and free recall of lists \cite{Franklin2015,Murdock1982}.

\subsection{Declarative Memory}
In \cite{Kelly2020hdm}, we implement a holographic declarative memory consisting of many individual holographic vectors. Each holographic vector represents a distinct concept, collectively serving as the basis vectors for the agent's conceptual space \cite{lieto2017conceptual}. Our model is able accounts for human performance on a wide range of tasks, including recall, probability judgement, and decision-making \cite{Kelly2020hdm}, as well as how humans learn the meaning and part-of-speech of words from experience \cite{Kelly2020indirect}.

\section{Validation}
\label{sec:validation}

We intend to validate the architecture by comparing to human data on standard memory and learning tasks, probability judgement tasks, and decision-making tasks. We will also test the architecture's ability to solve problems in environments starting with simple games that require agile adaptation, such as rock-paper-scissors and mazes, graduating slowly to more complex environments, e.g., video game worlds, all the while comparing to state-of-the-art statistical learning methods. Continual learning benchmarks such as Split MNIST, Fashion MNIST, and NotMNIST would also provide a useful starting point in evaluating the systems's performance when processing overlapping and non-overlapping (in terms of input distribution) sets/groups of patterns sequentially \cite{ororbia2019lifelong}.

\section{Conclusions and Future Research}
\label{sec:conclusion}

Humans are capable of (a) lifelong, continual learning and deep expertise, (b) single-trial learning and agile adaptation to dynamic environments, and (c) transfer learning across multiple tasks. Conversely, conventional neural networks struggle to replicate these human capabilities. 
Solving the problem of continual learning will aid us both in understanding the brain processes that facilitate learning, inference, and memory storage in human minds and in the development of intelligent agents that are better able to generalize to real world environments. 
Crucially, our proposed instantiation of the common model of cognition is grounded in and composed of neuro-cognitively plausible components, i.e., holographic memory, predictive processing circuits, and competitive learning. As a result, several promising research directions open up but among them is the application of this architecture to situations where the challenge of catastrophic interference is most prevalent -- reinforcement learning across infinite streams of tasks where knowledge retrieval, transfer, and composition are critical \cite{zhan2017scalable,khetarpal2018environments}.

\bibliographystyle{acm}
\bibliography{ref}

\end{document}